\documentclass[conference]{IEEEtran}
\usepackage{times}

\usepackage[numbers]{natbib}
\usepackage{multicol}
\usepackage[bookmarks=true]{hyperref}
\usepackage{graphicx}
\usepackage{amsmath} 

\usepackage{multirow}
\usepackage{textcomp}
\usepackage{gensymb}
\usepackage{float}
\usepackage{amssymb}
\usepackage{subcaption} 
\usepackage[ruled,linesnumbered]{algorithm2e}
\usepackage{algorithmic}
\usepackage[dvipsnames]{xcolor}
\usepackage{blkarray}
\newcommand{\etal}{\textit{et al}.}

\begin{document}
	
\title{\LARGE \bf A Unified Method for Solving Inverse, Forward, and Hybrid \\ Manipulator Dynamics using Factor Graphs}


\author{\authorblockN{Mandy Xie and Frank Dellaert}
	\authorblockA{School of Interactive Computing, 
		Georgia Institute of Technology\\
		Atlanta, Georgia 30332--0250, 
		Email: \{manxie,dellaert\}@gatech.edu}
}


%

\maketitle

\begin{abstract}
This paper describes a unified method solving for inverse, forward, and hybrid dynamics problems for robotic manipulators with either open kinematic chains or closed kinematic loops based on factor graphs. Manipulator dynamics is considered to be a well studied problem, and various different algorithms have been developed to solve each type of dynamics problem. However, they are not easily explained in a unified and intuitive way. In this paper, we introduce factor graphs as a unifying graphical language in which not only to solve all types of dynamics problems, but also explain the classical dynamics algorithms in a unified framework.
\end{abstract}

\IEEEpeerreviewmaketitle

\section{Introduction \& Related Work}
\begin{figure*}[!ht]
	\centering
	\includegraphics[scale=1]{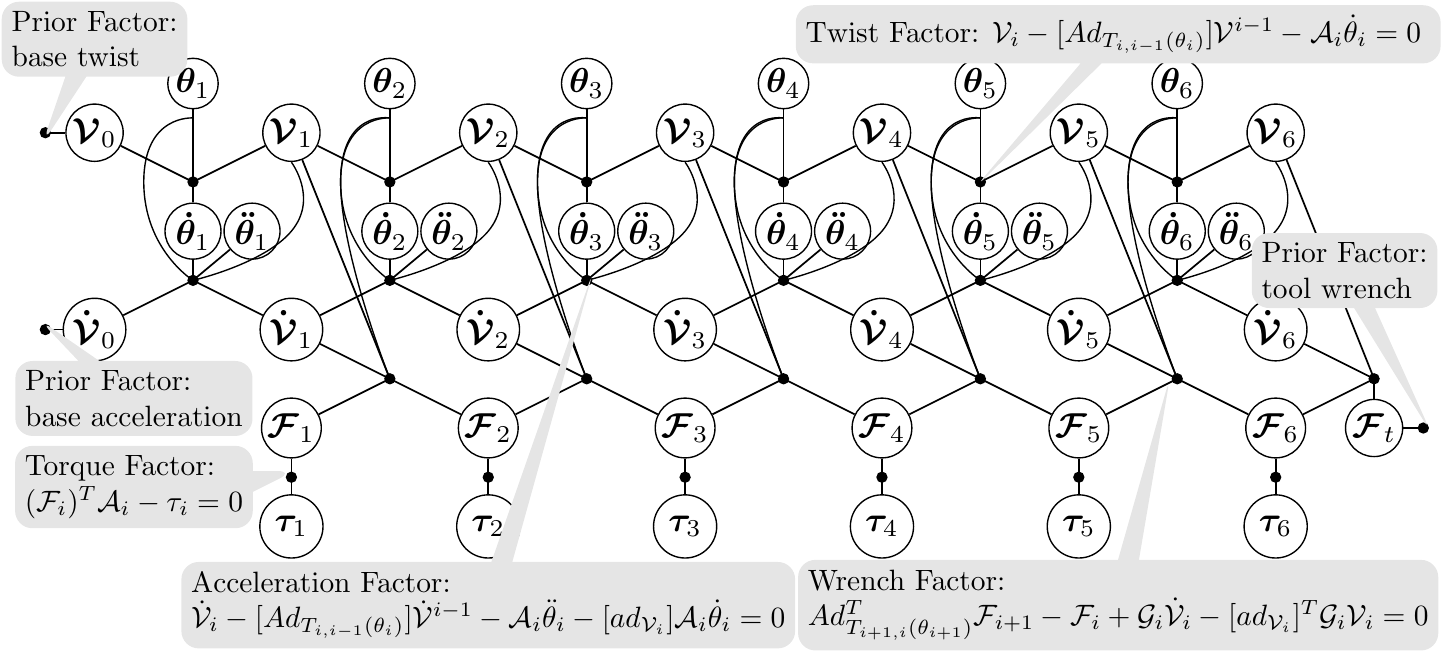}
	\caption{The Puma 560 dynamics factor graph, where black dots represent factors, and circles represent variables.}
	\label{fig:DFG-puma}
\end{figure*}

There are three main types of problems involved in the study of manipulator dynamics: inverse, forward, and hybrid problems. 
Inverse dynamics, which is used in control and motion planning, calculates the torques required at the joints to generate a desired trajectory of joint positions, velocities and accelerations. The Newton-Euler (N-E)~\cite{Craig09book_robotics} method is one of the key approaches in solving inverse dynamics problems since it results in very efficient recursive algorithms, such as RNEA~\cite{Luh80jdsmc_manipulator,Felis17jar_rbdl} and similar methods described in~\cite{Orin79jmb_NE} and~\cite{Stepanenko76jmb_dynamics}. 
Forward dynamics, which is primarily used in the simulation of robotic manipulators, determines joints accelerations with torques applied at the joints, again given joint positions and velocities. Algorithms based on the inertia matrix method include the Composite-Rigid-Body Algorithm (CRBA)~\cite{Walker82jdsmc_dynamics,Featherstone00icra_crba} and propagation methods such as the Articulated-Body Algorithm (ABA)~\cite{Featherstone83ijrr_aba}. 
Finally, hybrid dynamics problems are sometimes used to incorporate prescribed motions for manipulators, and works out the unknown forces and accelerations with given forces at some joints and accelerations at other joints. Since neither inverse nor forward dynamics algorithms can be directly applied, solutions typically combine elements from both inverse and forward methods, e.g., the articulated-body hybrid dynamics algorithm described in Section 9.2 of~\cite{Featherstone14book_dynamics}. 

The methods mentioned above do not apply for manipulators with kinematic loops. 
Because of the complexity caused by kinematic redundancy~\cite{Chiaverini08shr_redundancy}, actuation redundancy~\cite{Nakamura89tra_redundancy}, and uncertainty in constraint forces exerted by loop joints~\cite{Featherstone14book_dynamics}, more sophisticated and expensive algorithms are required to calculate their dynamics, which can be found in~\cite{Rodriguez86isrm_forward-dynamics,Nakamura89tra_redundancy} and  Chapter 8 of~\cite{Featherstone14book_dynamics}.
More sophisticated and expensive algorithms are required to solve both inverse and forward dynamics for parallel robots, which can be found in~\cite{Featherstone14book_dynamics,Rodriguez86isrm_forward-dynamics}.

There is rarely a single algorithm which can solve all three types of dynamics problems. Different algorithms have to be applied as described in Section 5.3, 6.2, 7.3 of~\cite{Featherstone14book_dynamics}. Rodriguez~\cite{Rodriguez87jra_forward-inverse-dynamics,Rodriguez86isrm_forward-dynamics} built a unified framework based on the concept of filtering and smoothing to solve both inverse and forward dynamics, and such method can also be applied to closed kinematic loops dynamics as claimed in~\cite{Rodriguez89tra_loop}. Different algorithms have to be designed and implemented, though they share a unified framework. Based on the work by Rodriguez et al.~\cite{Rodriguez91ijrr_soa}, Jain~\cite{Jain91jgcd_unified} analyzed various algorithms for serial chain dynamics in a unified formulation. Ascher et al.~\cite{Ascher97ijrr_forward-dynamics} unified the derivation of CRBA and ABA as two elimination methods which are used to solve forward dynamics.
Different algorithms have to be designed for each type~\cite{Featherstone14book_dynamics}, and they are not easy to be explained in an intuitive way. Rodriguez~\cite{Rodriguez87jra_forward-inverse-dynamics} used a random field estimation approach to solve both inverse and forward dynamics problems, which builds parallelism between the concepts of estimation used in Kalman filtering and smoothing theory and the dynamics problems. However, the parallelism is not straightforward and easy to visualize. Even though it is claimed in~\cite{Rodriguez86isrm_forward-dynamics} that the same method can be used to solve dynamics for manipulators involving closed kinematic loop, non-trivial modifications have to be made for this method to be applied.

In this paper we introduce factor graphs as a unifying language in which to explain the classical dynamics algorithms, and present new algorithms derived from the graph theory underpinning sparse linear systems. Our contributions are:
\begin{itemize}
\item a unified method which can solve inverse, forward and hybrid dynamics for either kinematic chains or loops;
\item a factor graph representation for dynamics problems, which is a insightful visualization of the underlying equations;
\item the discovery of new dynamics algorithms corresponding to different elimination orderings in those graphs.
\end{itemize}




Note that graphical models in general have been used before in robotic dynamics, e.g., Ting \etal~\cite{Ting06rss_identification} used Bayes networks to model system identification of rigid body parameters from noisy data. Factor graphs have also been used in walking robots by at least two different groups ~\cite{Hartley18icra_fg_estimation, Hartley18iros_hybrid, Wisth19ral_legged_fg}, but without explicit modeling of dynamical quantities as we do below.

We are also not the first ones to exploit ordering/permutations of matrices to improve performance: sparse linear algebra was already exploited very early on in \cite{Orlandea77asme_sparsity} and is still being re-discovered regularly, e.g. \cite{Nori17rcar_sparse}. Rather, the factor graph approach exposes a very general view on these problems and exposes elimination ordering in a much clearer way than typical sparse linear algebra methods, and opens up the possibility of adding non-linearities into the same framework. Finally, the eliminated directed acyclic graphs (DAGs) (Section \ref{sec:inverse}) point the way to automatically generating code corresponding to a topological ordering of the corresponding DAG.

\section{Review of Manipulator Dynamics}
Below we review the modern geometric view on framing robot dynamics, and follow the exposition and notation from the recent text by Lynch and Park~\cite{Lynch17book_robotics}. As convincingly argued in their introduction, this geometric view pioneered by Brockett~\cite{Brockett84mtns_robotics} and Murray et al.~\cite{Murray94book}, unlocks the powerful tools of modern differential geometry to reason about robot dynamics. It will also help below in describing the various dynamics algorithms in a concise graphical representation.
 
\newcommand{\I}{\mathcal{I}}
\newcommand{\w}{\omega}
\newcommand{\wdot}{\dot{\w}}

Traditionally, the Newton-Euler equations of motion for a rigid body moving in space subjects to external forces $f$ and torques $\tau$, can be expressed in body coordinates as (Equations 8.22 and 8.23 on page 242 in \cite{Lynch17book_robotics}),
\begin{align} 
\label{eq:rtd}
f_b &= m\dot{v}_b + \w_b \times mv_b \\
\label{eq:rrd}
\tau_b &= \I_b \wdot_b + \w_b \times \I_b \w_b
\end{align}
with $m$, $\I_b$, $v_b$, and $\w_b$ respectively the mass, inertia, linear and angular velocity expressed in body coordinate frame.

\newcommand{\V}{\mathcal{V}}
\newcommand{\Vdot}{\mathcal{\dot{V}}}
\newcommand{\G}{\mathcal{G}}
\newcommand{\F}{\mathcal{F}}
\newcommand{\q}{\theta}
\newcommand{\qdot}{\dot{\q}}
\newcommand{\qddot}{\ddot{\q}}
\newcommand{\Axis}{\mathcal{A}}

In the geometric view, we combine equations \eqref{eq:rtd} and \eqref{eq:rrd} to obtain an equation in terms of the six-dimensional body wrench $\F_b$ and body twist $\V_b$ quantities (Equation 8.40 on page 247 of \cite{Lynch17book_robotics}),
\begin{equation} \label{eq:crd}
    \F_b = \G_b \dot{\V}_b - [ad_{\V_b}]^T \G_b \V_b
\end{equation}
where the new quantities are defined as
$$
\V_b =  \left[ \begin{array}{c} \w_b \\ v_b \end{array} \right] \ \
\F_b =  \left[ \begin{array}{c} \tau_b \\ f_b \end{array} \right] $$
$$\G_b = \begin{bmatrix} \I_b & 0 \\ 0 & m I \end{bmatrix} \ \
[ad_{\V_b}] = \begin{bmatrix} [\w_b] & 0 \\ [v_b] & [\w_b] \end{bmatrix}
$$
Above $[\w_b]$ is the skew-symmetric matrix formed from $\w_b$, i.e.,
$$
[\w_b]=R^T \dot{R}
$$
with $R\in SO(3)$ the rotation associated with a link.

Closely following Section 8.3 in~\cite{Lynch17book_robotics}, applying this to the links of a serial manipulator and taking into account the constraints at the joints, we obtain four equations relating both link and joint quantities.
In particular, the twist and acceleration $\V_i$ and $\Vdot_i$ for the $i$-th link are expressed in a body-fixed coordinate frame rigidly attached to the link. The wrench transmitted through joint $i$ is denoted as $\F_i$, and $\G_i$ is the link's inertia matrix. Without loss of generality, below we assume all rotational joints, and we then have: 
\begin{align} 
\label{eq:twist}
    \V_i - [Ad_{T_{i,i-1}(\q_i)}]\V_{i-1} - \Axis_i\qdot_i &= 0 \\
\label{eq:accel}
    \Vdot_i - [Ad_{T_{i,i-1}(\q_i)}]\Vdot_{i-1} - \Axis_i\qddot_i - [ad_{\V_i}]\Axis_i\qdot_i &= 0  \\
\label{eq:wrench}
    Ad^T_{T_{i+1,i}(\q_{i+1})}\F_{i+1} - \F_i + \G_i\dot{\V}_i - [ad_{\V_i}]^T\G_i\V_i &= 0 \\
\label{eq:torque}
    \F_i^T\Axis_i - \tau_i &= 0
\end{align}
where $\Axis_i$ is the screw axis for joint $i$ (expressed in link $i$), and $Ad_{T_{i,i-1}(\q_i)}$ is the adjoint transformation associated with the transform $T_{i,i-1}$ between the links (a function of $\q_i$).

The four equations \ref{eq:twist}-\ref{eq:torque} express the dynamic constraints between link $i$ and link $i-1$ imposed by joint $i$: 
\eqref{eq:twist} describes the relationship between twist $\V_i$ and twist $\V_{i-1}$, where $\qdot_i$ is the angular velocity of joint $i$; 
\eqref{eq:accel} describes the constraint between acceleration $\Vdot_i$ of and acceleration $\Vdot_{i+1}$, which involves components due to joint acceleration $\qddot_i$ and the acceleration caused by rotation;
\eqref{eq:wrench} describes the balance between the wrench $\F_i$ through joint $i$ and the wrench $\F_{i+1}$ applied through joint $i+1$;
\eqref{eq:torque} describes that the torque applied at joint {i} equals to the projection of wrench $\F_i$ on the screw axis $\Axis_i$ corresponding to joint $i$.

Gravity is not considered above but can easily be accounted for. Lynch \& Park~\cite{Lynch17book_robotics} describe a standard "trick" that adds an extra acceleration term to the base. While clever, we have found it more intuitive to explicitly deal with gravity in our implementation, where we simply add a gravity term to the wrench equation \eqref{eq:wrench}, expressed in each link's body frame.

\section{A Factor Graph Approach}
\subsection{Factor Graphs}

A factor graph~\cite{Dellaert17fnt_fg} is a graphical model that can be used to describe the structure of sparse computational problems. It is used in constraint satisfaction~\cite{Seidel81ijcai,Freuder82jacm,Dechter87ai}, AI~\cite{Pearl88book,Shenoy86expert,Lauritzen88jrssb, Frey97ccc,Kschischang01it}, sparse linear algebra~\cite{George84laa,Gilbert93chapter,Heggernes96siam}, information theory~\cite{Tanner81it,Loeliger04spm}, combinatorial optimization~\cite{Bertele72book,Bertele72jmaa,Bertele73jct}, and even query theory~\cite{Beeri81stoc,Fagin82tods,Goodman82tods}.  A general theory specified in terms of algebraic semirings was also developed~\cite{Carre71jima}, and seminal work in theory proved essential computational complexity results~\cite{Lipton79nd} based on
the existence of separator theorems for certain classes of graphs
~\cite{Lipton79sep}.
Factor graphs have been successfully applied in other areas of robotics, such as SLAM~\cite{Kaess08tro,Kaess12ijrr,Dellaert17fnt_fg,Forster17tro}, state estimation in humanoids \cite{Hartley18iros_hybrid}, and (kinematic) motion planning~\cite{Dong16rss,Mukadam16icra,Mukadam17rss}.

\subsection{Dynamic Factor Graphs}

\begin{figure}
	\centering
	\includegraphics[scale=0.23]{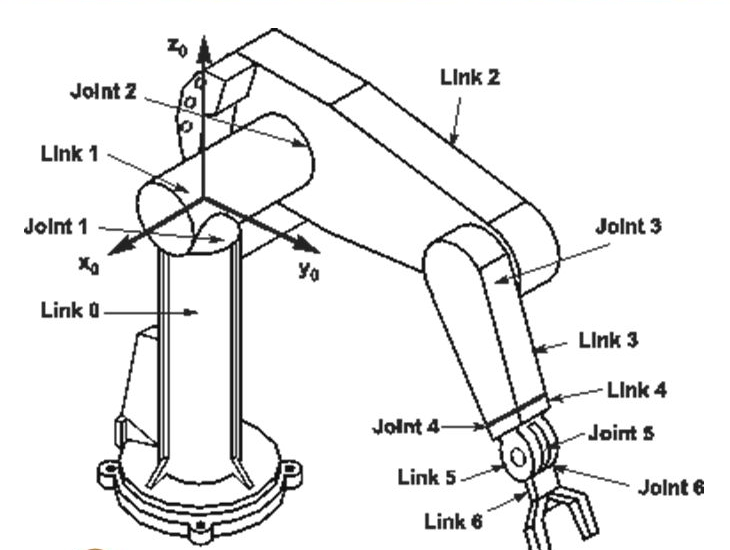}
	\caption{The PUMA 560 robot~\cite{Armstrong86icra_puma560}.}
	\label{fig:puma-robot}
\end{figure}

A key idea is that we can use a factor graph to represent the structure of the dynamics constraints \eqref{eq:twist}-\eqref{eq:torque} for a particular manipulator configuration. 
A factor graph consists of \emph{factors} and \emph{nodes}, where factors correspond to the dynamics constraints, and the nodes represent the variables in each equation.
Factors are only connected to the variable nodes that are featured in the corresponding dynamics constraint, revealing the sparsity structure of the system of dynamics equations.
Figure~\ref{fig:DFG-puma} illustrates this for the classical Puma 560 robot shown in Figure~\ref{fig:puma-robot}.
Variables including twists $\V_i$, accelerations $\Vdot_i$, wrenches $\F_i$, joint angles $\q_i$, joint velocities $\qdot_i$, joint accelerations $\qddot_i$, and torques $\tau_i$. 
The repetitive sparse structure of the 6R robot can be clearly observed.

The dynamics factor graph corresponding to all variables and constraints can be simplified and used to solve the different types of dynamics problems, i.e., inverse, forward, and hybrid problems. We show this in detail in the three following sections. 
However, due to space constraints, we use a three link RRR example for the remainder of this paper.

\begin{figure}
	\begin{center}
		\includegraphics[scale=0.93]{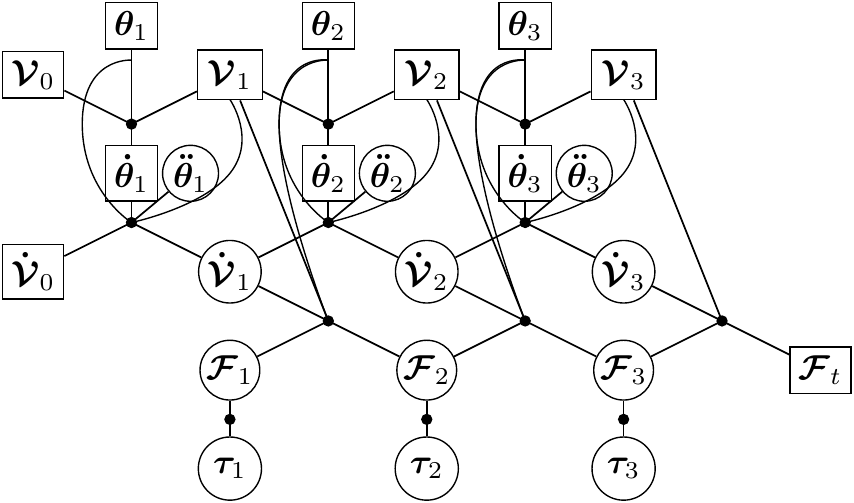}
	    \caption{Dynamic factor graph for a 3R robot, with \emph{known} variables shown as square nodes.}
		\label{fig:simple3R}
	\end{center}
\end{figure}

Also, in all three problems, we typically assume that the kinematic quantities $\q_i$ and $\qdot_i$ are known for all joints. 
This in turn allows us to solve for the twist variables $\V_{i}$ in advance.
In addition, we typically also assume that $\Vdot_{0}$ and the end-effector wrench $\F_{t}$ are given, as well.
In the language of graphical models it is common to denote \emph{known} variables as square nodes. This is illustrated
in Fig.~\ref{fig:simple3R} for an RRR robot, which will be the starting point for the sections below.

\subsection{Automatic Transcription into a Factor Graph}
The dynamic factor graphs for all results below are obtained automatically by converting Unified Robot Description Format (URDF) files programatically into an internal representation that works with the GTSAM factor graph library~\cite{Dellaert12tr, Dellaert17fnt_fg}. This process is relatively straightforward, and the code will be released in the public domain. 
In essence, a bipartite graph of joints and links is created, which is then transcribed into joint-specific, link-specific, and joint-link interaction factors, as shown in Figure \ref{fig:DFG-puma}. 
For parallel robots, which are not supported by the URDF format out of the box, we provide the ability to provided an amended URDF file with extra loop closures, or parse a \href{http://sdformat.org/}{Simulation Description Format (SDF)} file. Instructions for both formats are provided in the repository.
To help with reviewing, an anonymized version of the code is available on \href{https://anonymous.4open.science/r/3aa5296f-9aeb-498f-8909-c667741fc079/}{https://anonymous.4open.science} at \href{https://anonymous.4open.science/r/3aa5296f-9aeb-498f-8909-c667741fc079/}{this link}. 

\section{Inverse Dynamics}
\label{sec:inverse}

\begin{figure}
	\centering
	\begin{subfigure}[b]{0.2\textwidth}
		\centering
		\includegraphics[scale=0.93]{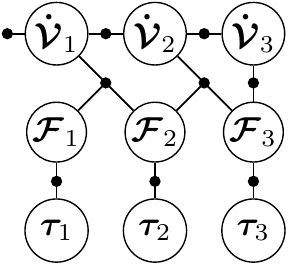}
		\caption{}
		\label{fig:IDFG-linear}
	\end{subfigure}
	\begin{subfigure}[b]{0.2\textwidth}
		\centering
		\includegraphics[scale=0.15]{figures/inverse_sparse_3R.png}
		\caption{}
		\label{fig:ID-sparse}
	\end{subfigure}
	\caption{(a) Simplified inverse dynamics factor graph (b) Block-sparse matrix corresponding to Fig.~\ref{fig:IDFG-linear}}
	\label{fig:IDFG-sparse}
\end{figure}

In inverse dynamics, we are seeking the required joint torques $\tau_i$ to realize the desired joint accelerations $\qddot_i$.
We can construct a simplified, less cluttered, \emph{inverse dynamics graph} by simply omitting all known nodes, although they remain as parameters in the factors they were connected to. For the RRR example, the resulting graph is shown in Fig.~\ref{fig:IDFG-linear}, corresponding to the 9 linear  constraints comprising the 3R inverse dynamics problem.

\subsection{Gaussian Elimination to a DAG}

The Gaussian elimination algorithm to solve this set of linear equations can be graphically understood as converting the factor graph in Fig.~\ref{fig:IDFG-linear} to a directed acyclic graph (DAG).
Just as a factor graph is the graphical embodiment of the dynamics system, the DAG reveals the sparsity structure resulting after Gaussian elimination of the dynamics equations.

Elimination proceeds one variable at a time, and expresses that variable in terms of variables that will be eliminated later. If more than one equation is involved, this will result in new equations, i.e., factors, that can lead to so-called \emph{fill-in} in the corresponding sparse system of linear equations.
This graphical "elimination game" was first developed in sparse linear algebra~\cite{George93book,Gilbert93chapter,Heggernes96siam},
but is also used in probabilistic inference, where the DAG represents a Bayes net~\cite{Pearl88book}, and in sensor fusion problems in robotics~\cite{Dellaert17fnt_fg}.

\begin{figure}
	\centering
	\begin{subfigure}[b]{0.2\textwidth}
		\centering
		\includegraphics[scale=0.93]{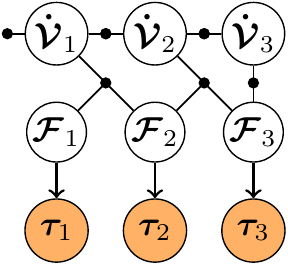}
		\caption{Eliminate all the $\tau$.}
		\label{fig:elimination1}
	\end{subfigure}
	\begin{subfigure}[b]{0.2\textwidth}
		\centering
		\includegraphics[scale=0.93]{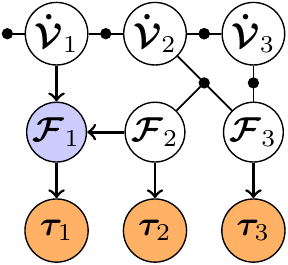}
		\caption{Eliminate $\F_1$.}
		\label{fig:elimination2}
	\end{subfigure}
	\begin{subfigure}[b]{0.2\textwidth}
		\centering
		\includegraphics[scale=0.93]{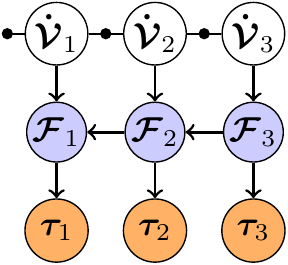}
		\caption{Eliminate all the $\F$.}
		\label{fig:elimination3}
	\end{subfigure}
	\begin{subfigure}[b]{0.2\textwidth}
		\centering
		\includegraphics[scale=0.93]{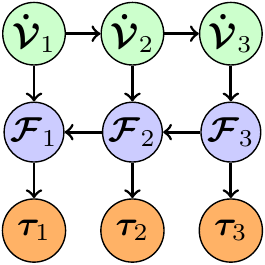}
		\caption{Eliminate all the $\Vdot$.}
		\label{fig:elimination4}
	\end{subfigure}
	\caption{Steps in the Elimination Algorithm.}
	\label{fig:IDFG-elimination}
\end{figure}

We illustrate the elimination process in the 3R case for a particular ordering in Fig.~\ref{fig:IDFG-elimination}. 
The elimination is performed in the order $\tau_3\dots\tau_1$, $\F_1\dots\F_3$, $\Vdot_3\dots\Vdot_1$.
Fig.~\ref{fig:elimination1} shows the result of first eliminating the torques $\tau_i$, where the arrows show that the torques $\tau_i$ only depend on the corresponding wrenches $\F_i$.
Fig.~\ref{fig:elimination2} shows the elimination of $\F_1$, which results in a dependence of $\F_1$ on $\Vdot_1$ and $\F_2$. 
After eliminating all the wrenches, we get the result as shown in Fig.~\ref{fig:elimination3}.
Finally, we eliminate the all the twist accelerations $\Vdot_3\dots\Vdot_1$, in that order. After completing all these elimination steps, the inverse dynamics factor graph in Fig.~\ref{fig:IDFG-linear} is thereby converted to the DAG as shown in Fig.~\ref{fig:elimination4}.

\subsection{Solving Symbolically}

\definecolor{myGreen}{RGB}{179, 255, 179}
\definecolor{myBlue}{RGB}{153, 153, 255}
\definecolor{myOrange}{RGB}{255, 153, 51}

\algsetup{indent=5em}
\begin{algorithm}
\caption{Inverse dynamics corresponding to Fig.~\ref{fig:elimination4}.\label{alg:RNEA3R}}
\colorbox{myGreen}{$\Vdot_1$} $= [Ad_{T_{1,0}(\q_1)}]\V_0 + [ad_{\V_1}]\Axis_1\qdot_1 + \Axis_1\qddot_1$ \;
\colorbox{myGreen}{$\Vdot_2$} $= [Ad_{T_{2,1}(\q_2)}]\colorbox{myGreen}{$\Vdot_1$} + [ad_{\V_2}]\Axis_2\qdot_2 + \Axis_2\qddot_2$ \;
\colorbox{myGreen}{$\Vdot_3$} $= [Ad_{T_{3,2}(\q_3)}]\colorbox{myGreen}{$\Vdot_2$} + [ad_{\V_3}]\Axis_3\qdot_3 + \Axis_3\qddot_3$ \;
\colorbox{myBlue}{$\F_3$} $= Ad^T_{T_{t,3}}\F_{t} + \G_3\colorbox{myGreen}{$\Vdot_3$} - [ad_{\V_3}]^T\G_3\V_3$ \;
\colorbox{myBlue}{$\F_2$} $= Ad^T_{T_{3,2}(\q_{3})}\colorbox{myBlue}{$\F_3$} + \G_2\colorbox{myGreen}{$\Vdot_2$} - [ad_{\V_2}]^T\G_2\V_2$ \;
\colorbox{myBlue}{$\F_1$} $= Ad^T_{T_{2,1}(\q_{2})}\colorbox{myBlue}{$\F_2$} + \G_1\colorbox{myGreen}{$\Vdot_1$} - [ad_{\V_1}]^T\G_1\V_1$ \;
\colorbox{myOrange}{$\tau_1$} $= \colorbox{myBlue}{$\F_1$}^T\Axis_1$ \;
\colorbox{myOrange}{$\tau_2$} $= \colorbox{myBlue}{$\F_2$}^T\Axis_2$ \;
\colorbox{myOrange}{$\tau_3$} $= \colorbox{myBlue}{$\F_3$}^T\Axis_3$ \;
\end{algorithm}

Elimination can be done numerically or symbolically. 
A symbolic elimination step can be very simple if only one equation is involved, or rather complicated if many equations are involved. Hence, it it matters which variables are eliminated first. For example, eliminating $\tau_3$ above is simply a matter of rewriting \eqref{eq:torque}
$$\F_3^T\Axis_3 - \tau_3 = 0$$
\vspace{-2mm}
as 
\vspace{-2mm}
$$\tau_3 = \F_3^T\Axis_3$$
However, if one were to eliminate $\F_2$ in Fig.~\ref{fig:IDFG-linear} first, it would involve three constraints (the number of factors attached to $\F_2$) and five variables, leading to two new constraints in those variables. That complexity will propagate to the rest of the graph, i.e., creating (symbolic) fill-in.

After elimination, back-substitution \emph{in reverse elimination order} solves for the values of all intermediate quantities and the desired torques. For the example ordering, this corresponding to the chosen order above first computes the 6-dimensional twists accelerations $\Vdot_i$, link wrenches $\F_i$, and then the scalar torques.
The back-substitution sequence corresponding to Fig.~\ref{fig:elimination4} can be written down as an \emph{algorithm}. The resulting algorithm for the 3R case and the chosen ordering is shown above as Algorithm \ref{alg:RNEA3R}.

The above \emph{exactly} matches the forward-backward path used by the recursive Newton-Euler algorithm (RNEA)~\cite{Luh80jdsmc_manipulator}.
The resulting DAG can be viewed as a graphical representation of RNEA, where the green and blue/orange colors resp. correspond to the forward path and the backward path.

\subsection{Solving Numerically}

However, we can also construct these factor graphs on the fly, for arbitrary robot configurations, and solve them numerically.
The symbolic elimination leads to very fast hard-coded dynamics algorithms, but have to be re-derived for every configuration. 
The numerical approach is exactly what underlies sparse linear algebra solvers, and can be extended to deal with over-constrained least-squares problems, in which the elimination algorithm corresponds to QR factorization.

For an arbitrary configuration, the numerical elimination (in arbitrary order) corresponds to a blocked Gaussian elimination where most blocks are $6\times6$, except where the (scalar) torques $\tau_i$ are concerned. 
For example, Fig~\ref{fig:ID-sparse} shows the sparse block-matrix corresponding to the simplified 3R inverse dynamics graph in Fig.~\ref{fig:IDFG-linear}. Every row in that matrix is associated with a factor in the graph, and every column with a variable node.
After elimination, back-substitution 
corresponding to the chosen order above first computes the 6-dimensional twists accelerations $\Vdot_i$, link wrenches $\F_i$, and then the scalar torques.

\begin{table}
	\caption{Numerical inverse dynamics for a PUMA 560}
	\label{tab:ID-EO}
	\begin{center}
		\begin{tabular}{| c | c |}
			\hline 
			Elimination Method & Average Time($\mu s$)\\
			\hline
			RNEA & 26.6 \\
			\hline
			RNEA in RBDL & 20.2 \\
			\hline
			COLAMD & 11.0 \\ 
			\hline
			ND  & 11.7 \\   
			\hline
		\end{tabular}
	\end{center}
\end{table} 

As already hinted at above, the cost of elimination on a factor graph can vary dramatically for different variable orderings, since different orderings lead to different DAG topologies. The amount of fill-in in turn affects the computational complexity of the elimination and back-substitution algorithms~\cite{Dellaert17fnt_fg}. Unfortunately, finding an optimal ordering is NP-complete and already intractable for a 6R robot, so ordering heuristics are used. Table~\ref{tab:ID-EO} shows that the RNEA ordering as discussed in Lynch \& Park is apparently outperformed by the custom implementation in RBDL~\cite{Felis17jar_rbdl}. This is in turn outperformed by COLAMD~\cite{Amestoy96siam} and nested dissection (ND)~\cite{George73siam}, which are two state of the art sparse linear algebra ordering heuristics. 

The reported results were obtained using GTSAM~\cite{Dellaert12tr}, a general factor graph solver used extensively in the robotics community. However, we make no claim that these results in any way come close to dedicated dynamics solvers, and they are intended to indicate relative performance rather than claiming SOA absolute performance. GTSAM is a very general library which is optimized towards much larger problems. In addition, once an ordering is chosen we should be able to perform symbolic elimination only once, rather than doing it every time as in our results. As alluded to above, in future work we plan to automatically generate code for particular robot topologies, which we hypothesize will match and possibly exceed existing solvers when using non-intuitive but computationally more advantageous elimination orderings.

\subsection{The Space of all Inverse Dynamics Algorithms}
Given all of the above, it is clear that the underlying graph theory formalizes the existence of an entire space of possible inverse dynamics algorithms: for every of the (intractably many) possible variable orderings, we have both a numerical and a symbolic variant. In theory, given enough time, we can exhaustively search all orderings for a given configuration and the generate a hard-coded algorithm that is optimal for that configuration.

\begin{figure}
	\centering
	\begin{subfigure}[b]{0.2\textwidth}
		\centering
		\includegraphics[scale=0.90]{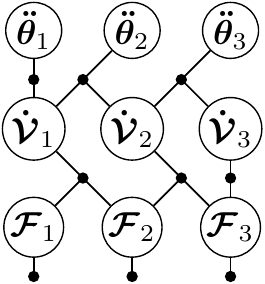}
		\caption{}
		\label{fig:FDFG-linear}
	\end{subfigure}
	\begin{subfigure}[b]{0.2\textwidth}
		\centering
		\includegraphics[scale=0.15]{figures/forward_sparse_3R.png}
		\caption{}
		\label{fig:FD-sparse}
	\end{subfigure}
	\caption{(a) Simplified forward dynamics graph and (b) corresponding block-sparse matrix.}
\end{figure}

\section{Forward Dynamics}
\begin{figure*}[!h]
	\centering
	\begin{subfigure}[b]{0.24\textwidth}
		\centering
		\includegraphics[scale=0.93]{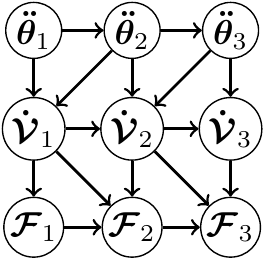}
		\caption{CRBA ordering.}
		\label{fig:FDBN-CRBA}
	\end{subfigure}
	\begin{subfigure}[b]{0.24\textwidth}
		\centering
		\includegraphics[scale=0.93]{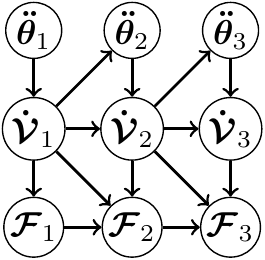}
		\caption{ABA ordering.}
		\label{fig:FDBN-ABA}
	\end{subfigure}
	\begin{subfigure}[b]{0.24\textwidth}
		\centering
		\includegraphics[scale=0.93]{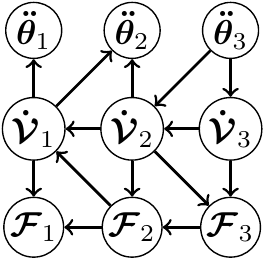}
		\caption{COLAMD ordering.}
		\label{fig:FDBN-COLAMD}
	\end{subfigure}
	\begin{subfigure}[b]{0.24\textwidth}
		\centering
		\includegraphics[scale=0.93]{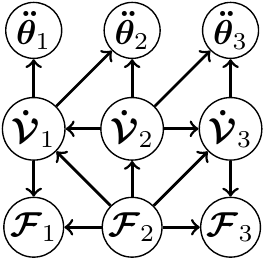}
		\caption{ND ordering.}
		\label{fig:FDBN-METIS}
	\end{subfigure}
	\caption{DAGs generated by solving forward dynamics factor graph with different variable elimination orderings.}
	\label{fig:FDBN}
\end{figure*}

In traditional expositions the forward dynamics problem cannot be solved as neatly as the relatively easy inverse dynamics problem.
In the forward case we are seeking the joint accelerations $\qddot$ when given $\q$, $\qdot$, and $\tau$. Similar to the inverse dynamics factor graph, we can simplify forward dynamics factor graph as shown in Fig.~\ref{fig:FDFG-linear}. 

\subsection{Solving Symbolically}

In very much the same spirit as our work, Ascher et al.~\cite{Ascher97ijrr_forward-dynamics} showed that two of the most widely used forward  algorithms, CRBA~\cite{Featherstone00icra_crba} and ABA~\cite{Featherstone83ijrr_aba}, can be explained as two different elimination methods to solve the same linear system. 

In our framework, CRBA and ABA can additionally be visualized as two different DAGs resulting from solving forward dynamics factor graph with two different elimination orders shown in Fig.~\ref{fig:FDBN-CRBA} and Fig.~\ref{fig:FDBN-ABA}.
CRBA can be explained as first eliminating all the wrenches $\F_i$, then eliminating all the accelerations $\Vdot_i$, and lastly eliminating all the angular accelerations $\qddot_i$.
%
In contrast, in ABA we alternate eliminating the wrenches $\F_i$, accelerations $\Vdot_i$ and angular accelerations $\qddot_i$ for $i\in n\dots1$.
%
The resulting DAGs can be viewed as graphical representations of CRBA and ABA, and for a given robot configuration, a custom back-substitution program can be written out in reverse elimination order. 

The forward dynamics problem is usually more complicated than the inverse dynamics problem, and hence there are more edges in the corresponding DAGs. However, better elimination orders lead to better algorithms in terms of operation counts.  
Indeed, different elimination orderings result in different fill-in, and an ordering with minimum fill-in minimizes the cost of the elimination algorithm!\cite{Dellaert17fnt_fg}.
For example, from Figures \ref{fig:FDBN-CRBA} and \ref{fig:FDBN-ABA} we can see that there are more edges in the CRBA DAG than in the ABA DAG, which means more computation is required using CRBA. This was already remarked upon by Ascher et al.~\cite{Ascher97ijrr_forward-dynamics}, but in the graphical framework we can tell this directly by observing the DAG.   


\subsection{Solving Numerically}

\begin{table}
	\caption{Numerical forward dynamics for a PUMA 560}
	\label{tab:FD-EO}
	\begin{center}
		\begin{tabular}{| c | c | c |}
			\hline 
			Elimination Method & Average Time($\mu s$)\\
			\hline
			CRBA & 51.8 \\ 
			\hline
			ABA & 25.5 \\
			\hline
			COLAMD & 11.2 \\ 
			\hline
			ND & 12.1 \\   
			\hline
		\end{tabular}
	\end{center}
\end{table}

Forward dynamics for arbitrary robot configurations can be solved by constructing the factor graphs on the fly and solving them numerically, using a sparse solver such as GTSAM~\cite{Dellaert12tr}. Similarly to the inverse case, we can use different variable ordering heuristics to explore the computational complexity of each scheme. In Table~\ref{tab:FD-EO} we report on four different orderings, applied to the PUMA 650 configuration, and we can see that ABA indeed outperforms CRBA in this case. However, the two sparse linear algebra ordering heuristics COLAMD and ND outperform both, by yet another factor of two or more.

\section{Hybrid Dynamics}

In hybrid dynamics problems, either $\qddot_i$ or $\tau_i(t)$ at each joint are given, and the task is to obtain the unknown accelerations and torques. To solve this problem, Featherstone introduced a hybrid dynamics algorithm in Section 9.2 of~\cite{Featherstone14book_dynamics}, where the set of joints for with the torques given but accelerations are unknown is denoted as "forward-dynamics joints", and the others are denoted as "inverse-dynamics joints". 

\begin{figure}
	\centering
	\includegraphics[scale=1]{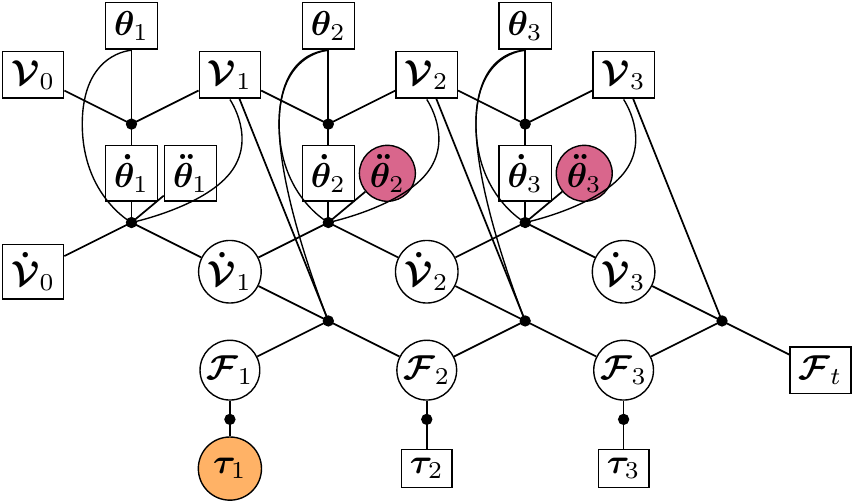}
	\caption{Hybrid dynamics factor graph for a 3R robot.}
	\label{fig:HDFG}
\end{figure}


\begin{figure}
	\centering
	\begin{subfigure}[b]{0.2\textwidth}
		\centering
	    \includegraphics[scale=0.93]{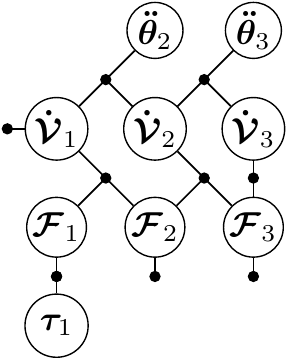}
	    \caption{}
	    \label{fig:HDFG-linear}
	\end{subfigure}
	\begin{subfigure}[b]{0.2\textwidth}
		\centering
		\includegraphics[scale=0.17]{figures/hybrid_sparse_3R.png}
		\caption{}
		\label{fig:hD-sparse}
	\end{subfigure}
	\caption{(a) Simplified hybrid dynamics graph and (b) corresponding block-sparse.}
	\label{fig:HDFG-sparse}
\end{figure}

\subsection{Featherstone's method}
Solving the hybrid dynamics using Featherstone's algorithm can be illustrated with factor graphs using a simple 3R example. Fig.~\ref{fig:HDFG} shows a factor graph for the case when $\tau_1$ is unknown while $\qddot_1$ is given, and additionally $\qddot_2$ and $\qddot_3$ are unknown while $\tau_2$ and $\tau_3$ are given. 
For this combination of given and unknown values, the hybrid dynamics factor graph can be simplified to Fig.~\ref{fig:HDFG-linear}. 


\begin{itemize}
    \item \textbf{Inverse Dynamics (Zero Acceleration Torques)}: Set $\qddot_i$ as known variables, where the values are the desired accelerations for $i = 1$, and the values are zeros for $i = 2$ and $3$; Set $\tau_i$ as known variables for $i = 2$ and $3$, where the values are given; Calculate $\tau_1$ with the inverse dynamics factor graph.
	\item \textbf{Forward Dynamics}: Set $\tau_i$ as known, where the values are from zero acceleration torques for $i = 1$, and the values for  $i = 2$ and $3$ are as given; Calculate $\qddot_i$ for $i = 2$ and $3$ with the forward dynamics factor graph.
	\item \textbf{Inverse Dynamics}: Set $\qddot_i$ to be known variables, where the values are as given for $i = 1$, and the values are from the last step for $i = 2$ to $3$; Calculate $\tau_1$ with the inverse dynamics factor graph.
\end{itemize}

%
%
%
	
\subsection{Using elimination in a Factor Graph}

\begin{figure}
	\centering
	\begin{subfigure}[b]{0.2\textwidth}
		\centering
		\includegraphics[scale=1]{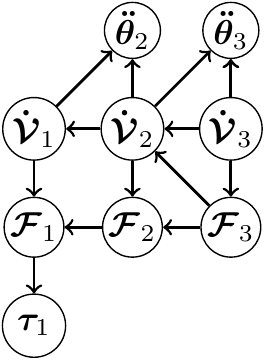}
		\caption{MD ordering.}
		\label{fig:HDBN-MD}
	\end{subfigure}
	\begin{subfigure}[b]{0.2\textwidth}
		\centering
		\includegraphics[scale=1]{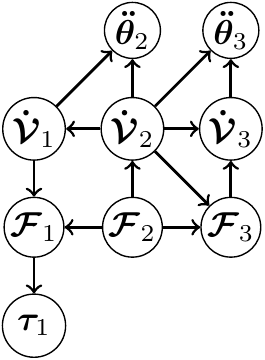}
		\caption{Custom ordering.}
		\label{fig:HDBN-CUSTOM}
	\end{subfigure}
	\caption{DAGs from solving hybrid dynamics factor graph.}
	\label{fig:HD-linear}
\end{figure}
It is not necessary to solve inverse and forward dynamics multiple times, because both forward-dynamics joints and inverse-dynamics joints can be solved simultaneously with the hybrid dynamics factor graph in Fig.~\ref{fig:HDFG}.
Using an elimination variable ordering generated by a minimum degree (MD) heuristic shown in Table~\ref{HD-EO}, the resulting DAG obtained is shown in Fig.~\ref{fig:HDBN-MD}.
For good measure, we also eliminated the factor graph with another, manually created elimination order listed as "CUSTOM" in Table~\ref{HD-EO}, and show the corresponding DAG in Fig.~\ref{fig:HDBN-CUSTOM}. The two DAGs are slightly different, but both have the same number of directed edges and hence might be suspected to have the same computational complexity. 

However, especially for hybrid problems like this, being sophisticated about variable ordering and the possible resulting parallelism could yield large dividends.
An important step forward in the understanding and analysis of variable
elimination on graphs was the discovery of \emph{clique trees}, that make the inherent parallelism in the elimination algorithm explicit~\cite{Liu89siam,Tarjan84siam,Blair93chapter}. A clique tree or directed Bayes tree~\cite{Kaess12ijrr} can be constructed from the DAG to guide parallel execution. 
The complexity of the numerical elimination step depends on the \emph{tree width}, i.e., the size of the largest clique in the tree.
For example, the variable ordering associated with the DAG in Fig.~\ref{fig:HDBN-CUSTOM} splits the graph on the clique formed by $\F_2$ and $\Vdot_2$. By taking advantage of this parallelism, we can solve this type of hybrid dynamics problem more efficiently.
\emph{Nested dissection} (ND) algorithms~\cite{George73siam} try to exploit this
by recursively partitioning the graph and returning a \emph{post-fix}
notation of the partitioning tree as the ordering. 
\begin{table}
	\caption{Elimination Orders for Hybrid Dynamics}
	\label{HD-EO}
	\begin{center}
		\begin{tabular}{| c | c |}
			\hline 
			Elimination Method & Elimination Order \\   
			\hline
			MD
			& $t_1, a_2, a_3,\Vdot_3, \F_1, \Vdot_1, \F_2, \Vdot_2, \F_3$ \\
			\hline
			CUSTOM
			& $t_1, a_2, a_3, \F_1, \Vdot_1, \Vdot_3, \F_3, \Vdot_2, \F_2$ \\
			\hline
		\end{tabular}
	\end{center}
\end{table}

\section{Dynamics with Closed Kinematic Loops}

\begin{figure}
	\centering
	\includegraphics[scale=0.5]{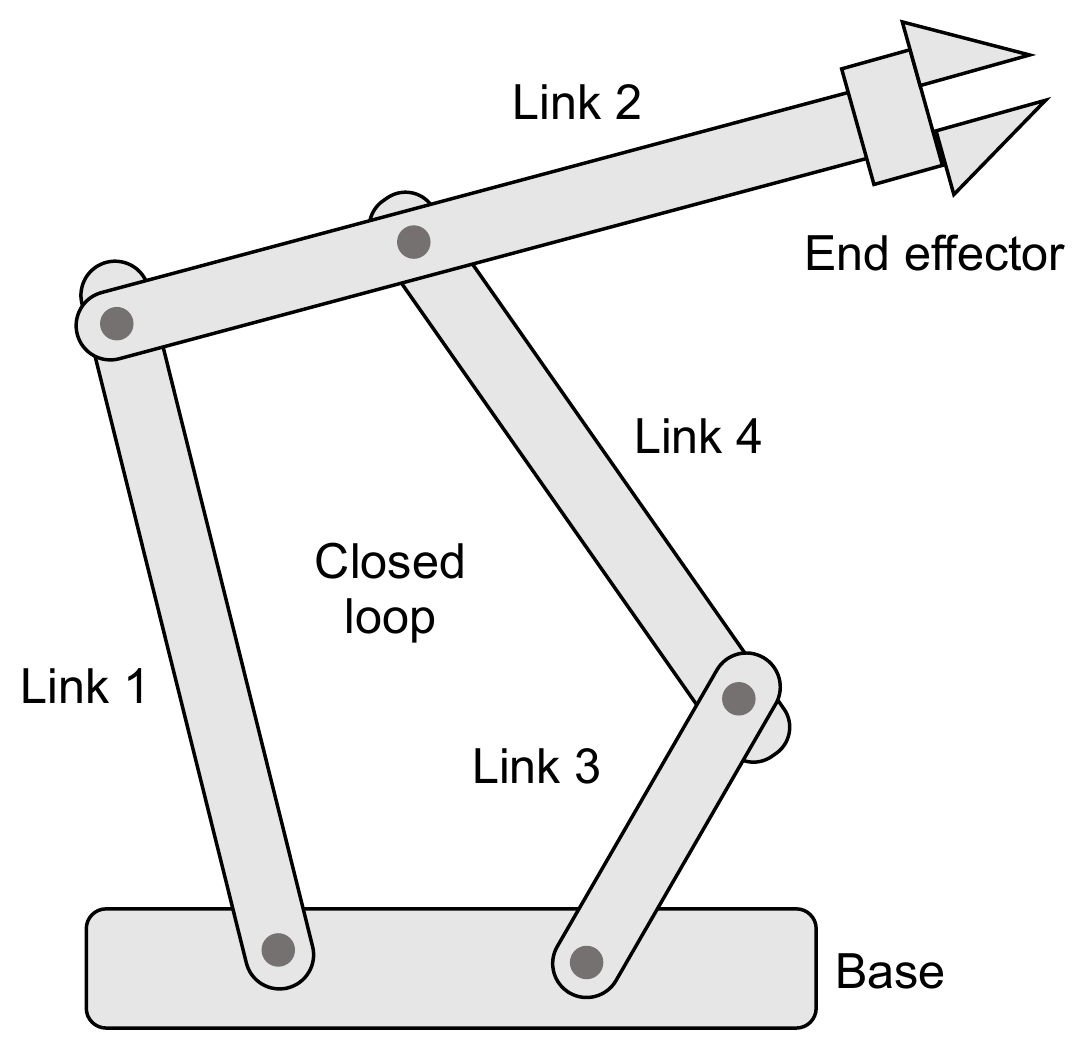}
	\caption{Five-bar parallel manipulator (adapted from ~\cite{5bar}). The bottom 2 joints are actuated but other joints move freely.}
	\label{fig:5-bar}
\end{figure}

\begin{figure*}
	\centering
	\includegraphics[scale=1.1]{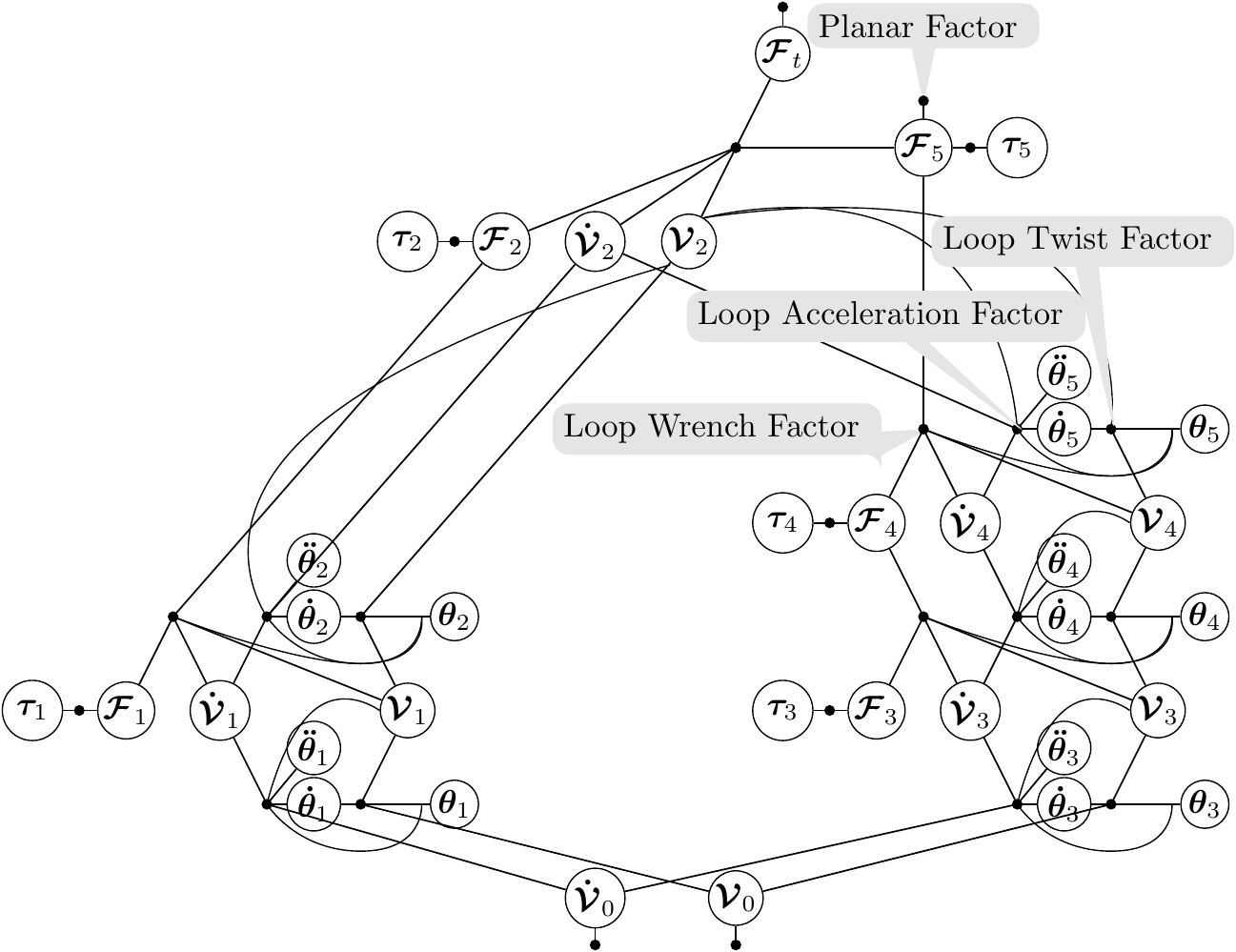}
	\caption{Five-bar parallel manipulator dynamics factor graph, where black dots represent variable constraints, namely factors, and circles represent variable nodes.}
	\label{fig:5BarDFG}
\end{figure*}

As discussed in Chapter 8 of~\cite{Featherstone14book_dynamics}, for inverse dynamics, if a manipulator with kinematic loop is redundantly actuated (the number of actuated joints is greater than the degree of motion freedom) there are infinitely many values of torque $\tau$ that produce the same angular acceleration $\qddot$. If a unique solution is required, one can either add more constraints or apply an optimality criterion, which can be done by adding extra factors to the graph. For example, minimum torque factors make the solution unique by choosing the minimum torque values. For forward dynamics, if a manipulator with kinematic loop is overconstrained (for example, any system containing planar kinematic loops), the constraint forces exerted by loop joints are underdetermined. We can convert this overconstrained system to be properly constrained if possible, for example, by replacing the original loop joint with a joint that imposes less constraints. With factor graphs, this can be done by adding a planar factor at the loop joint which reduces the number of unknown constraint forces so it can be properly solved.

We use a five-bar parallel manipulator as shown in Fig.~\ref{fig:5-bar} to illustrate how to solve kinematic loops with factor graphs. This manipulator is properly actuated, since only joints 1 and 2 are actuated, and other joints are free to rotate. Joint 5 closes the loop. Link 0 represent the base link, which is fixed, and the end-effector is attached to link 2. Since the kinematic loop in this manipulator is planar, we add a planar factor to the dynamics factor graph at the loop joint as shown in Fig.~\ref{fig:5BarDFG}, where the planar factor is shown as a unary factor associated with wrench $\F_5$, which is the unknown constraint wrench exerted by the loop joint.
With the closed loop dynamics factor graph, we can solve inverse, forward and hybrid dynamics problems by specifying which variables are known and which are unknown. We can solve this factor graph with any elimination ordering, and the resulting DAGs can be taken as graphical representations for different algorithms to solve closed-loop problems.

\section{Discussion}
In this paper, we represent manipulator dynamics as factor graph and solve for inverse, forward, and hybrid problems. Using factor graphs as a graphical language gives us not only a unified method to solve all types of dynamics problems, but also an insightful visualization of the underlying mathematical formulations. Exploiting different elimination orders of solving the factor graph unlocks powerful tools to illustrate classical algorithms and derive novel algorithms which could be applied to solve certain types of dynamics problems efficiently. 

As we discussed Section \ref{sec:inverse}, the reported timing results are in no way intended to compete with dedicated dynamics solvers, but rather indicated relative performance. In future work we plan to automatically generate code for particular robot topologies, which we hypothesize will match and possibly exceed existing solvers when using non-intuitive but computationally more advantageous elimination orderings. We are also aware that in comparing high performance code controlling for cache effects and memory architecture in general is important, as done by Neuman \etal~\cite{Neuman19iros_benchmarking}.

In future work, we hope to apply these findings to perform kinodynamic motion planning in the style of GPMP2~\cite{Mukadam18ijrr} and STEAP~\cite{Mukadam18ar}, which successfully applied incremental inference in factor graphs~\cite{Kaess12ijrr, Dellaert17fnt_fg} to kinematic motion planning problems. In addition, it would be very interesting to use the factor-graph-based representation of dynamics to perform state estimation for dynamically balanced robots, in the spirit of Hartley \etal~\cite{Hartley18icra_fg_estimation, Hartley18iros_hybrid} and Wisth \etal~\cite{Wisth19ral_legged_fg}.


\newpage
\bibliographystyle{plain}
\bibliography{references.bib}

\begin{thebibliography}{10}

\bibitem{Amestoy96siam}
P.R. Amestoy, T.~Davis, and I.S. Duff.
\newblock An approximate minimum degree ordering algorithm.
\newblock {\em SIAM Journal on Matrix Analysis and Applications},
  17(4):886--905, 1996.

\bibitem{Armstrong86icra_puma560}
B.~Armstrong, O.~Khatib, and J.~Burdick.
\newblock The explicit dynamic model and inertial parameters of the puma 560
  arm.
\newblock In {\em Proceedings. 1986 IEEE international conference on robotics
  and automation}, volume~3, pages 510--518. IEEE, 1986.

\bibitem{Ascher97ijrr_forward-dynamics}
U.~M. Ascher, P.~K. Dinesh, and B.~P. Cloutier.
\newblock Forward dynamics, elimination methods, and formulation stiffness in
  robot simulation.
\newblock {\em The International Journal of Robotics Research}, 16(6):749--758,
  1997.

\bibitem{Beeri81stoc}
C.~Beeri, R.~Fagin, D.~Maier, A.~Mendelzon, J.~Ullman, and M.~Yannakakis.
\newblock Properties of acyclic database schemes.
\newblock In {\em ACM Symp. on Theory of Computing (STOC)}, pages 355--362, New
  York, NY, USA, 1981. ACM Press.

\bibitem{Bertele72jmaa}
U.~Bertele and F.~Brioschi.
\newblock On the theory of the elimination process.
\newblock {\em J. Math. Anal. Appl.}, (1):48--57, July.

\bibitem{Bertele72book}
U.~Bertele and F.~Brioschi.
\newblock {\em Nonserial Dynamic Programming}.
\newblock Academic Press, 1972.

\bibitem{Bertele73jct}
U.~Bertele and F.~Brioschi.
\newblock On nonserial dynamic programming.
\newblock {\em J. Combinatorial Theory}, 14:137--148, 1973.

\bibitem{Blair93chapter}
J.R.S. Blair and B.W. Peyton.
\newblock An introduction to chordal graphs and clique trees.
\newblock In George et~al. \cite{_George93edited}, pages 1--27.

\bibitem{Brockett84mtns_robotics}
W.~Roger Brockett.
\newblock Robotic manipulators and the product of exponentials formula.
\newblock In {\em Mathematical theory of networks and systems}, pages 120--129.
  Springer, 1984.

\bibitem{Carre71jima}
B.~A. Carr\'{e}.
\newblock An algebra for network routing problems.
\newblock {\em J. Inst. Math. Appl.}, 7:273--294, 1971.

\bibitem{Chiaverini08shr_redundancy}
S.~Chiaverini, G.~Oriolo, and I.~D. Walker.
\newblock Kinematically redundant manipulators.
\newblock {\em Springer handbook of robotics}, pages 245--268, 2008.

\bibitem{Craig09book_robotics}
J.~J Craig.
\newblock {\em Introduction to robotics: mechanics and control, 3/E}.
\newblock Pearson Education India, 2009.

\bibitem{Dechter87ai}
R.~Dechter and J.~Pearl.
\newblock Network-based heuristics for constraint-satisfaction problems.
\newblock {\em Artificial Intelligence}, 34(1):1--38, December 1987.

\bibitem{Dellaert12tr}
F.~Dellaert.
\newblock Factor graphs and {GTSAM}: A hands-on introduction.
\newblock Technical Report GT-RIM-CP\&R-2012-002, Georgia Institute of
  Technology, 2012.

\bibitem{Dellaert17fnt_fg}
F.~Dellaert and M.~Kaess.
\newblock Factor graphs for robot perception.
\newblock {\em Foundations and Trends in Robotics}, 6(1-2):1--139, 2017.

\bibitem{Dong16rss}
J.~Dong, M.~Mukadam, F.~Dellaert, and B~Boots.
\newblock Motion planning as probabilistic inference using {G}aussian processes
  and factor graphs.
\newblock In {\em Robotics: Science and Systems (RSS)}, 2016.

\bibitem{Fagin82tods}
R.~Fagin, A.O. Mendelzon, and J.D. Ullman.
\newblock A simplied universal relation assumption and its properties.
\newblock {\em ACM Trans. Database Syst.}, 7(3):343--360, 1982.

\bibitem{Featherstone83ijrr_aba}
R.~Featherstone.
\newblock The calculation of robot dynamics using articulated-body inertias.
\newblock {\em The International Journal of Robotics Research}, 2(1):13--30,
  1983.

\bibitem{Featherstone14book_dynamics}
R.~Featherstone.
\newblock {\em Rigid body dynamics algorithms}.
\newblock Springer, 2014.

\bibitem{Featherstone00icra_crba}
R.~Featherstone and D.~E. Orin.
\newblock Robot dynamics: equations and algorithms.
\newblock In {\em Proceedings 2000 ICRA. Millennium Conference. IEEE
  International Conference on Robotics and Automation. Symposia Proceedings
  (Cat. No. 00CH37065)}, volume~1, pages 826--834. IEEE, 2000.

\bibitem{Felis17jar_rbdl}
M.~L. Felis.
\newblock {RBDL}: {An} efficient rigid-body dynamics library using recursive
  algorithms.
\newblock {\em Autonomous Robots}, 41(2):495--511, 2017.

\bibitem{Forster17tro}
C.~Forster, L.~Carlone, F.~Dellaert, and D.~Scaramuzza.
\newblock On-manifold preintegration for real-time visual-inertial odometry.
\newblock {\em {IEEE} Trans. Robotics}, 2017.

\bibitem{Freuder82jacm}
Eugene~C. Freuder.
\newblock A sufficient condition for backtrack-free search.
\newblock {\em J. ACM}, 29(1):24--32, 1982.

\bibitem{Frey97ccc}
B.J. Frey, F.R. Kschischang, H.-A. Loeliger, and N.~Wiberg.
\newblock Factor graphs and algorithms.
\newblock In {\em Proc. 35th Allerton Conf. Communications, Control, and
  Computing}, pages 666--680, September 1997.

\bibitem{George73siam}
A.~George.
\newblock Nested dissection of a regular finite element mesh.
\newblock {\em SIAM Journal on Numerical Analysis}, 10(2):345--363, April 1973.

\bibitem{George84laa}
A.~George, J.~Liu, and Ng~E.
\newblock Row-ordering schemes for sparse {Givens} transformations. {I.
  Bipartite} graph model.
\newblock {\em Linear Algebra Appl}, 61:55--81, 1984.

\bibitem{George93book}
J.A. George, J.R. Gilbert, and J.W-H. Liu, editors.
\newblock {\em Graph Theory and Sparse Matrix Computations}, volume~56 of {\em
  IMA Volumes in Mathematics and its Applications}.
\newblock Springer-Verlag, 1993.

\bibitem{_George93edited}
J.A. George, J.R. Gilbert, and J.W-H. Liu, editors.
\newblock {\em Graph Theory and Sparse Matrix Computations}, volume~56 of {\em
  IMA Volumes in Mathematics and its Applications}.
\newblock Springer-Verlag, New York, 1993.

\bibitem{Gilbert93chapter}
J.R. Gilbert and E.G. Ng.
\newblock Predicting structure in nonsymmetric sparse matrix factorizations.
\newblock In George et~al. \cite{_George93edited}.

\bibitem{Goodman82tods}
N.~Goodman and O.~Shmueli.
\newblock Tree queries: a simple class of relational queries.
\newblock {\em ACM Trans. Database Syst.}, 7(4):653--677, 1982.

\bibitem{Hartley18iros_hybrid}
R.~{Hartley}, M.~G. {Jadidi}, L.~{Gan}, J.~{Huang}, J.~W. {Grizzle}, and R.~M.
  {Eustice}.
\newblock Hybrid contact preintegration for visual-inertial-contact state
  estimation using factor graphs.
\newblock In {\em 2018 IEEE/RSJ International Conference on Intelligent Robots
  and Systems (IROS)}, pages 3783--3790, Oct 2018.

\bibitem{Hartley18icra_fg_estimation}
R.~{Hartley}, J.~{Mangelson}, L.~{Gan}, M.~{Ghaffari Jadidi}, J.~M. {Walls},
  R.~M. {Eustice}, and J.~W. {Grizzle}.
\newblock Legged robot state-estimation through combined forward kinematic and
  preintegrated contact factors.
\newblock In {\em 2018 IEEE International Conference on Robotics and Automation
  (ICRA)}, pages 4422--4429, May 2018.

\bibitem{Heggernes96siam}
P.~Heggernes and P.~Matstoms.
\newblock Finding good column orderings for sparse {QR} factorization.
\newblock In {\em Second SIAM Conference on Sparse Matrices}, 1996.

\bibitem{Jain91jgcd_unified}
A.~Jain.
\newblock Unified formulation of dynamics for serial rigid multibody systems.
\newblock {\em Journal of Guidance, Control, and Dynamics}, 14(3):531--542,
  1991.

\bibitem{Kaess12ijrr}
M.~Kaess, H.~Johannsson, R.~Roberts, V.~Ila, J.~Leonard, and F.~Dellaert.
\newblock {iSAM2}: Incremental smoothing and mapping using the {B}ayes tree.
\newblock {\em Intl. J. of Robotics Research}, 31:217--236, Feb 2012.

\bibitem{Kaess08tro}
M.~Kaess, A.~Ranganathan, and F.~Dellaert.
\newblock {iSAM}: Incremental smoothing and mapping.
\newblock {\em {IEEE} Trans. Robotics}, 24(6):1365--1378, Dec 2008.

\bibitem{Kschischang01it}
F.R. Kschischang, B.J. Frey, and H-A. Loeliger.
\newblock Factor graphs and the sum-product algorithm.
\newblock {\em {IEEE} Trans. Inform. Theory}, 47(2), February 2001.

\bibitem{Lauritzen88jrssb}
S.~L. Lauritzen and D.~J. Spiegelhalter.
\newblock Local computations with probabilities on graphical structures and
  their application to expert systems.
\newblock {\em Journal of the Royal Statistical Society. Series B
  (Methodological)}, 50(2):157--224, 1988.

\bibitem{Lipton79nd}
R.J. Lipton, D.J. Rose, and R.E. Tarjan.
\newblock Generalized nested dissection.
\newblock {\em SIAM Journal on Applied Mathematics}, 16(2):346--358, 1979.

\bibitem{Lipton79sep}
R.J. Lipton and R.E. Tarjan.
\newblock A separator theorem for planar graphs.
\newblock {\em SIAM Journal on Applied Mathematics}, 36(2):177--189, April
  1979.

\bibitem{Liu89siam}
J.~W-H. Liu and A.~Mirzaian.
\newblock A linear reordering algorithm for parallel pivoting of chordal
  graphs.
\newblock {\em SIAM J. Disc. Math.}, 2:lOO--107, 1989.

\bibitem{Loeliger04spm}
H.-A. Loeliger.
\newblock An introduction to factor graphs.
\newblock {\em IEEE Signal Processing Magazine}, pages 28--41, January 2004.

\bibitem{Luh80jdsmc_manipulator}
J.~Y.~S. Luh, M.~W. Walker, and R.~P. Paul.
\newblock On-line computational scheme for mechanical manipulators.
\newblock {\em Journal of Dynamic Systems, Measurement, and Control},
  102(2):69--76, 1980.

\bibitem{Lynch17book_robotics}
K.~M Lynch and F.~C Park.
\newblock {\em Modern Robotics}.
\newblock Cambridge University Press, 2017.

\bibitem{Mukadam17rss}
M.~Mukadam, J.~Dong, F.~Dellaert, and B.~Boots.
\newblock Simultaneous trajectory estimation and planning via probabilistic
  inference.
\newblock In {\em Robotics: Science and Systems (RSS)}, 2017.

\bibitem{Mukadam18ar}
M.~Mukadam, J.~Dong, F.~Dellaert, and B.~Boots.
\newblock Steap: simultaneous trajectory estimation and planning.
\newblock {\em Autonomous Robots}, pages 1--20, 2018.

\bibitem{Mukadam18ijrr}
M.~Mukadam, J.~Dong, X.~Yan, F.~Dellaert, and B.~Boots.
\newblock Continuous-time gaussian process motion planning via probabilistic
  inference.
\newblock {\em Intl. J. of Robotics Research}, 37(11):1319--1340, 2018.

\bibitem{Mukadam16icra}
M.~Mukadam, X.~Yan, and B.~Boots.
\newblock Gaussian process motion planning.
\newblock In {\em IEEE Intl. Conf. on Robotics and Automation (ICRA)}, 2016.

\bibitem{Murray94book}
R.M. Murray, Z.~Li, and S.~Sastry.
\newblock {\em A Mathematical Introduction to Robotic Manipulation}.
\newblock CRC Press, 1994.

\bibitem{Nakamura89tra_redundancy}
Y.~Nakamura and M.~Ghodoussi.
\newblock Dynamics computation of closed-link robot mechanisms with
  nonredundant and redundant actuators.
\newblock {\em IEEE Transactions on Robotics and Automation}, 5(3):294--302,
  1989.

\bibitem{Neuman19iros_benchmarking}
S.~M. {Neuman}, T.~{Koolen}, J.~{Drean}, J.~E. {Miller}, and S.~{Devadas}.
\newblock Benchmarking and workload analysis of robot dynamics algorithms.
\newblock In {\em 2019 IEEE/RSJ International Conference on Intelligent Robots
  and Systems (IROS)}, pages 5235--5242, Nov 2019.

\bibitem{Nori17rcar_sparse}
F.~{Nori}.
\newblock Inverse, forward and other dynamic computations computationally
  optimized with sparse matrix factorizations.
\newblock In {\em 2017 IEEE International Conference on Real-time Computing and
  Robotics (RCAR)}, pages 371--377, July 2017.

\bibitem{Orin79jmb_NE}
D.~E. Orin, R.~McGhee, M.~Vukobratovi, and G.~Hartoch.
\newblock Kinematic and kinetic analysis of open-chain linkages utilizing
  newton-euler methods.
\newblock {\em Mathematical Biosciences}, 43(1-2):107--130, 1979.

\bibitem{Orlandea77asme_sparsity}
N~Orlandea and DA~Calahan.
\newblock A sparsity-oriented approach to the design of mechanical systems.
\newblock {\em Problem Analysis in Science and Engineering}, pages 361--389,
  1977.

\bibitem{Pearl88book}
J.~Pearl.
\newblock {\em Probabilistic Reasoning in Intelligent Systems: Networks of
  Plausible Inference}.
\newblock Morgan Kaufmann, 1988.

\bibitem{5bar}
Reddit Robotics.
\newblock
  \href{https://www.reddit.com/r/robotics/comments/6fzwoo/in_the_fivebarlink_parallel_link_robot_joints/}{5-bar
  parallel manipulator}.

\bibitem{Rodriguez86isrm_forward-dynamics}
G.~Rodriguez.
\newblock Recursive forward dynamics for two robot arms in a closed chain based
  on kalman filtering and bryson-frazier smoothing.
\newblock In {\em Proceedings of the International Symposium on Robot
  Manipulators on Recent trends in robotics: modeling, control and education},
  pages 85--93. Elsevier North-Holland, Inc., 1986.

\bibitem{Rodriguez87jra_forward-inverse-dynamics}
G.~Rodriguez.
\newblock Kalman filtering, smoothing, and recursive robot arm forward and
  inverse dynamics.
\newblock {\em IEEE Journal on Robotics and Automation}, 3(6):624--639, 1987.

\bibitem{Rodriguez89tra_loop}
G.~Rodriguez.
\newblock Recursive forward dynamics for multiple robot arms moving a common
  task object.
\newblock {\em IEEE Transactions on Robotics and Automation}, 5(4):510--521,
  1989.

\bibitem{Rodriguez91ijrr_soa}
G.~Rodriguez, A.~Jain, and K.~Kreutz-Delgado.
\newblock A spatial operator algebra for manipulator modeling and control.
\newblock {\em The International Journal of Robotics Research}, 10(4):371--381,
  1991.

\bibitem{Seidel81ijcai}
R.~Seidel.
\newblock A new method for solving constraint satisfaction problems.
\newblock In {\em Intl. Joint Conf. on AI (IJCAI)}, pages 338--342, 1981.

\bibitem{Shenoy86expert}
P.~P. Shenoy and G.~Shafer.
\newblock Propagating belief functions using local computations,.
\newblock {\em IEEE Expert}, 1(3):43--52, Fall 1986.

\bibitem{Stepanenko76jmb_dynamics}
Y.~Stepanenko and M.~Vukobratovi.
\newblock Dynamics of articulated open-chain active mechanisms.
\newblock {\em Mathematical Biosciences}, 28(1-2):137--170, 1976.

\bibitem{Tanner81it}
R.~Tanner.
\newblock A recursive approach to low complexity codes.
\newblock {\em {IEEE} Trans. Inform. Theory}, 27(5):533--547, Spetember 1981.

\bibitem{Tarjan84siam}
R.E. Tarjan and M.~Yannakakis.
\newblock Simple linear-time algorithms to test chordality of graphs, test
  acyclicity of hypergraphs and selectively reduce acyclic hypergraphs.
\newblock {\em SIAM J. Comput.}, 13(3):566--579, 1984.

\bibitem{Ting06rss_identification}
J-A Ting, M.~Mistry, J.~Peters, S.~Schaal, and J.~Nakanishi.
\newblock A bayesian approach to nonlinear parameter identification for rigid
  body dynamics.
\newblock In {\em Robotics: Science and Systems}, pages 32--39. Philadelphia,
  USA, 2006.

\bibitem{Walker82jdsmc_dynamics}
M.~W. Walker and D.~E. Orin.
\newblock Efficient dynamic computer simulation of robotic mechanisms.
\newblock {\em Journal of Dynamic Systems, Measurement, and Control},
  104(3):205--211, 1982.

\bibitem{Wisth19ral_legged_fg}
D.~{Wisth}, M.~{Camurri}, and M.~{Fallon}.
\newblock Robust legged robot state estimation using factor graph optimization.
\newblock {\em IEEE Robotics and Automation Letters}, 4(4):4507--4514, Oct
  2019.

\end{thebibliography}

\end{document}